\documentclass[letterpaper]{article} 
\usepackage{aaai2026}  
\usepackage{times}  
\usepackage{helvet}  
\usepackage{courier}  
\usepackage[hyphens]{url}  
\usepackage{graphicx} 
\usepackage{natbib}  
\usepackage{caption} 
\usepackage{algorithm}
\usepackage{algorithmic}

\usepackage{newfloat}
\usepackage{listings}
\DeclareCaptionStyle{ruled}{labelfont=normalfont,labelsep=colon,strut=off} 
\floatstyle{ruled}
\newfloat{listing}{tb}{lst}{}
\floatname{listing}{Listing}
\usepackage{amsmath}
\usepackage{amssymb}
\usepackage{booktabs}
\usepackage{multirow}
\usepackage{longtable}
\usepackage{makecell}
\usepackage{subcaption}

\nocopyright 

\title{CoCAViT: Compact Vision Transformer with Robust Global Coordination}
\author{
    Xuyang Wang,
    Lingjuan Miao,
    Zhiqiang Zhou
}
\affiliations{
    \textsuperscript{}Beijing Institute of Technology\\
}

\begin{document}

\maketitle

\begin{abstract}
 In recent years, large-scale visual backbones have demonstrated remarkable capabilities in learning general-purpose features from images via extensive pre-training. Concurrently, many efficient architectures have emerged that have performance comparable to that of larger models on in-domain benchmarks. However, we observe that for smaller models, the performance drop on out-of-distribution (OOD) data is disproportionately larger, indicating a deficiency in the generalization performance of existing efficient models. To address this, we identify key architectural bottlenecks and inappropriate design choices that contribute to this issue, retaining robustness for smaller models. To restore the global field of pure window attention, we further introduce a Coordinator-patch Cross Attention (CoCA) mechanism, featuring dynamic, domain-aware global tokens that enhance local-global feature modeling and adaptively capture robust patterns across domains with minimal computational overhead. Integrating these advancements, we present CoCAViT, a novel visual backbone designed for robust real-time visual representation. Extensive experiments empirically validate our design. At a resolution of 224×224, CoCAViT-28M achieves 84.0\% top-1 accuracy on ImageNet-1K, with significant gains on multiple OOD benchmarks, compared to competing models. It also attains 52.2 mAP on COCO object detection and 51.3 mIOU on ADE20K semantic segmentation, while maintaining low latency.
\end{abstract}

\section{Introduction}

Visual backbones have been playing a key role in a wide array of visual tasks, serving as the foundational encoders that extract general-purpose features from visual data \cite{he2016deep,xie2017aggregated,howard2017mobilenets}. As vision models scale up, powerful backbones with massive parameters and extensive pretraining have emerged across vision tasks \cite{radford2021learning,touvron2021training,bao2021beit,he2022masked}. Transformers, in particular, have surpassed traditional convolutional neural networks (CNNs) in representational capacity and flexibility, excelling in large-scale scenarios through mechanisms like dense self-attention \cite{NIPS2017_3f5ee243}. Models such as V-MoE \cite{riquelme2021scaling}, Swin Transformer v2 \cite{liu2022swin}, and EVA \cite{fang2023eva} highlight this shift, achieving state-of-the-art results on benchmarks like ImageNet-1K \cite{russakovsky2015imagenet}.

\begin{figure}[t]
\centering
\includegraphics[width=1\columnwidth]{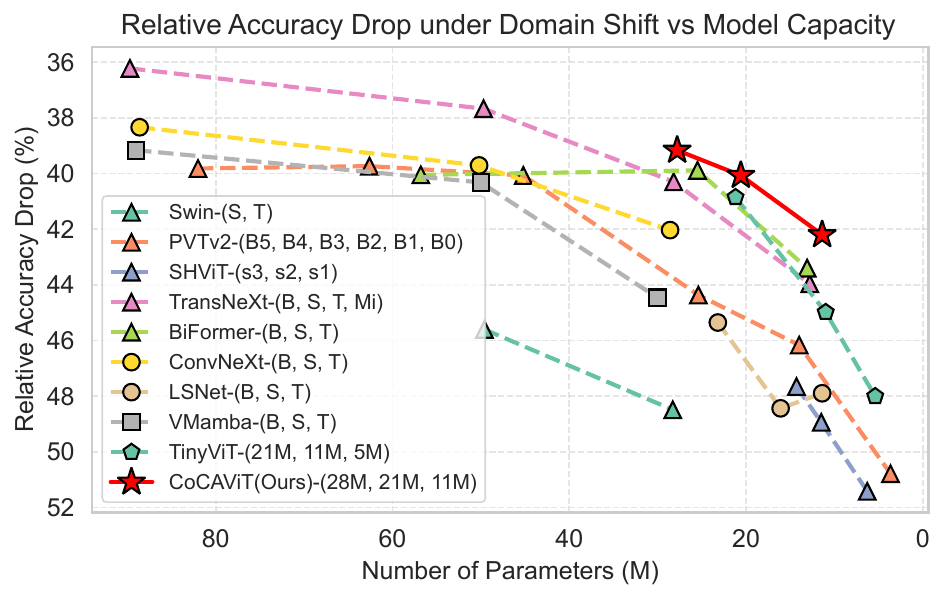} 
\caption{Relative accuracy drop under domain shift vs model capacity. We evaluate various vision architectures on OOD test set ImageNet-R, measuring the relative performance degradation compared to ImageNet-1K validation accuracy. Different markers represent different architectural families (triangles for Transformers, circles for CNNs, etc.). Our CoCAViT demonstrates superior robustness across model scales, particularly excelling in smaller parameter regimes where existing state-of-the-art methods suffer disproportionate performance drops.}
\label{fig1}
\end{figure}

However, the quadratic complexity of self-attention in Transformers poses severe challenges for real-time deployment on resource-constrained devices. This has spurred the development of efficient visual backbones, such as MobileViT \cite{mehta2021mobilevit}, LeViT \cite{graham2021levit}, and EfficientViT \cite{liu2023efficientvit}, which prioritize reduced complexity and inference speed. While these compact designs enable practical applications, they often compromise on representational depth, leading to vulnerabilities in handling complex feature distributions and distribution shifts. A pressing question thus arises: \textit{How can we design a compact visual backbone that is both sufficiently expressive for general-purpose representation while ensuring efficiency for real-time use?}

We observe that efficient models, despite competitive in-domain performance on benchmarks like ImageNet-1K, exhibit pronounced fragility under distribution shifts. To quantify this, we evaluate various architectures pretrained on ImageNet-1K, measuring relative accuracy drops on out-of-distribution test sets. As illustrated in Figure 1, compact models suffer steeper performance degradation—often over 45\% relative to in-distribution (ID) accuracy—compared to larger counterparts. This vulnerability arises from inherent architectural trade-offs: (1) aggressive scaling, such as diminished head dimensions and shallower depths, reduces attention diversity and hierarchical abstraction, fostering overfitting to training distributions; (2) reliance on local attention (e.g., window-based multi-head self attention in Swin Transformer \cite{liu2021swin} or sparse variants in EfficientViT) limits global receptive fields, with partial remedies like window shifting introducing inefficiencies and incomplete context integration. These limitations amplify sensitivity to shifts, hindering reliable deployment in dynamic real-world scenarios. This robustness-efficiency gap highlights the urgent need for architectural innovations that can maintain strong generalization capabilities while preserving computational efficiency.

In this work, we address these issues by exploring real-time visual backbones optimized for robust global coordination capabilities. 

\begin{itemize}
    \item We begin by identifying key architectural bottlenecks in efficient transformers through empirical analysis and propose principled design choices that enhance robustness while preserving efficiency. Specifically, we adopt a hybrid architecture with wider CNN stages for robust low-level feature extraction and deeper transformer stages with optimized head dimensions. We also introduce progressive MLP expansion ratios that allocate computational capacity more effectively across network depth.
    
    \item Our core innovation is the Coordinated Cross-Attention (CoCA) mechanism, which addresses the fundamental limitation of window-based attention—the lack of global context. CoCA introduces a sparse set of learnable coordinators that serve as dynamic global anchors, establishing bidirectional information flow through Gated Global Cross-Attention (GGCA) and Global-Coordinated Window Attention (GCWA). This design enables robust global coordination with linear complexity, effectively bridging window silos while maintaining computational efficiency.
    
    \item We introduce CoCAViT, a compact vision transformer with robust global coordination capabilities. In addition to the aforementioned key innovations, the model incorporates a Global Semantic Token Generator that produces domain-robust coordinators through parallel channel and spatial attention modules, enhanced by anchor loss regularization to ensure semantic consistency and diversity across coordinators. To enable hierarchical global coordination without prohibitive computational costs, we develop an attention-based Token Merging mechanism that compresses coordinators across stages while preserving their most semantically relevant information through cross-attention selection.
\end{itemize}

\section{Related Work}

\subsubsection{Efficient Vision Transformers.}
With the deepening research in Vision Transformers \cite{dosovitskiy2020image}, efficient variants like MobileViT \cite{mehta2021mobilevit} and EfficientViT \cite{liu2023efficientvit}, incorporate optimizations such as sparse attention, hierarchical structures, and reduced token counts, achieving comparable or even superior performance to larger models on in-domain tasks while drastically lowering latency.
Currently, hybrid architectures that combine CNN and ViT elements are recognized as an effective way to balance the strengths of both paradigms. These models result in superior performance in efficient settings \cite{li2023uniformer, graham2021levit}. Our work follows this hybrid architecture style, mitigating the lack of inductive biases in pure Transformers (e.g., locality and shift-invariance) while leveraging the large receptive fields of self-attention for enhanced global modeling.

\subsubsection{Local Attention Mechanisms.}
Local attention mechanisms have emerged as a key strategy to mitigate the quadratic complexity of vanilla self-attention in Vision Transformers, particularly in compact models. Swin Transformer \cite{liu2021swin} exemplifies this through its shifted window attention, which confines computations to non-overlapping windows and uses cyclic shifts to enable cross-window communication, achieving efficient hierarchical modeling. Similar approaches, such as BiFormer \cite{zhu2023biformer} with bi-level routing attention, dynamically sparsify attention maps to focus on relevant tokens, reducing overhead while approximating global context. However, these methods can still suffer from limited receptive fields and sensitivity to domain shifts due to their reliance on predefined or locally constrained sparsity patterns. Our CoCAViT draws inspiration from these sparse designs, particularly the shifted window paradigm in Swin, but extends it with coordinator-driven cross-attention that provides explicit, adaptive global coordination—enabling freer information flow across regions without the rigidity of shifts, thus enhancing robustness in distribution-shifted scenarios.

\begin{figure*}[t]
\centering
\includegraphics{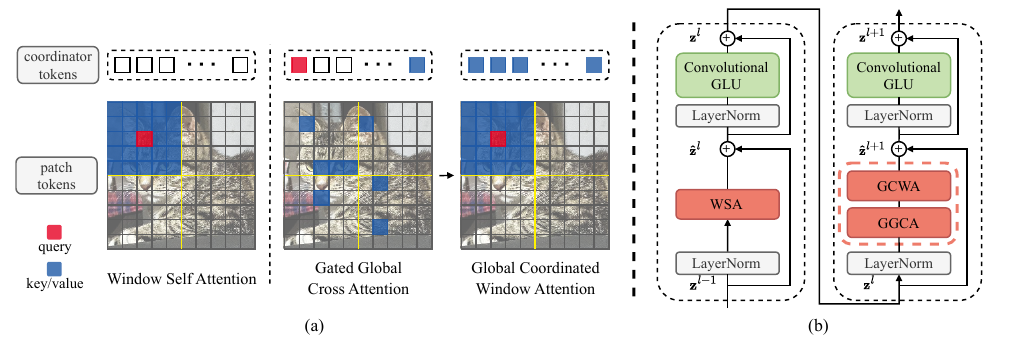} 
\caption{(a) Illustration and comparison of 3 adopted attentions: Window Self Attention (WSA), Gated Global Cross Attention (GGCA), and Global Coordinated Window Attention (GCWA). (b) Two successive CoCA Blocks.}
\label{fig2}
\end{figure*}

\subsubsection{Robustness in Vision Backbones.}
Enhancing robustness in vision backbones under distribution shifts is critical for reliable deployment. While domain generalization techniques—such as data augmentation \cite{zhou2021domain}, feature alignment \cite{li2018deep}, or domain-adversarial strategies \cite{ganin2016domain}—help handle unseen domains, broader robustness often arises from knowledge distillation \cite{hinton2015distilling, touvron2021training}, self-supervised learning \cite{chen2020simple, he2020momentum, caron2021emerging}, and architectural innovations like hybrid designs. Despite these, many overlook limitations in efficient architectures, such as restricted receptive fields, leading to suboptimal performance under constraints. In contrast, CoCAViT focuses on structural innovation for better robustness without multi-domain training emphasis.

\section{Methodology}
Designing efficient vision backbones for domain robustness poses unique challenges that are not fully addressed by current lightweight architectures. In this section, we analyze limitations in existing small-scale vision transformers from perspectives of local attention generalization and capacity-limited token interaction, using these insights to address bottlenecks through targeted architectural reconfiguration for enhanced robustness.

\subsection{Architectural Reconfiguration for Robustness}
\subsubsection{Inductive Biases at Early Stage.}
Incorporating convolutional neural network (CNN) layers in the initial stage of a hybrid backbone introduces crucial inductive biases, such as locality and translation equivariance, absent in pure Transformer architectures. These biases act as regularization, favoring domain-robust low-level patterns like edges and textures that show greater invariance across distributions than high-level semantics \cite{xiao2020noise}. Empirical evidence from LeViT \cite{graham2021levit} shows accelerated convergence in small Transformers via hybrid designs, with reduced gradient variance for reliable pretraining.

In our implementation, we use MBConv blocks from MobileNetV3 \cite{howard2019searching} for the first stage, downsampling inputs to H/4 × W/4. This enhances early-stage robustness through localized processing, provides a stable base for global coordination, and supports hardware acceleration.

\subsubsection{Wider CNN, Deeper Transformer.}
Compact hybrid backbones often face trade-offs where narrow early channels limit feature diversity and shallow Transformer depths hinder global abstraction, exacerbating fragility under distribution shifts in resource-constrained settings. To mitigate this, we adopt the \textit{wider CNN, deeper Transformer} reconfiguration optimizing for robust representation.

In the early CNN stage, we prioritize wider channels to enrich shallow feature extraction, capturing diverse low-level patterns such as orientations and textures that prove resilient across distributions. This draws from empirical findings in Wu et al. \cite{wu2019wider}, where shallow-and-wider CNN designs achieve comparable or superior performance to deeper variants with minimal computational overhead. Our experiments show broader layers enable smoother optimization, reducing initialization sensitivity and boosting stable gradients for robustness in small models during pretraining.

For the subsequent Transformer stages, we increase depth while decreasing head dimensions (without reducing head count), maintaining multi-head diversity at lower quadratic costs. Inspired by TransNeXt \cite{shi2024transnext}, which shows that thinner and deeper Transformers enhance robustness in efficient regimes, our design fosters progressive long-range dependencies and invariant feature abstraction, mitigating shifts via stronger gradient flow in OOD testing. This paradigm underpins global coordination token initialization, yielding stability and performance gains at iso-parameters.

\subsubsection{Progressive MLP Expansion Ratio.}
Feed-forward networks (FFNs) in Transformers, controlled by MLP expansion ratio, are key for capacity allocation. In compact models, static or early-biased ratios cause inefficient allocation, redundant later computations, and overfitting vulnerability under shifts. We introduce a progressive (decreasing) MLP ratio across stages, tuning expressivity adaptively for robust performance with minimal overhead.

Drawing from PVTv2's early-heavy ratios [8,8,4,4] \cite{wang2022pvt} and BiFormer's uniform low ratios [3,3,3,3] \cite{zhu2023biformer}, our scheme gradually reduces ratios, aligning the first stage with the MBConv expand ratio. This boosts early non-linear mixing for diverse, invariant representations, while later curbs reduce redundancy, aligning with efficient approximation principles \cite{jacot2018neural} and parsimony to dodge noise sensitivity. By prioritizing higher ratios upfront, we amplify robust pattern extraction with minimal parameter bloat, further reducing overall compute and parameters compared to uniform designs—while preserving expressive capacity, as validated in efficiency benchmarks.

\subsection{Coordinators-Patch Cross Attention Mechanism}

Local patterns limit receptive fields, hampering long-range dependencies and global context—vital for domain generalization. To address this limitation, we propose Coordinator-Patch Cross Attention (CoCA), bridging local-global modeling efficiently. CoCA adds learnable global coordinators as bottlenecks, aggregating robust features spatially and redistributing to boost local attention. Unlike shifted windows \cite{liu2021swin} using temporal shifts for connectivity, CoCA offers explicit cross-attention coordination, delivering better generalization with negligible overhead.

\begin{figure}[t]
\centering
\includegraphics[width=1\columnwidth]{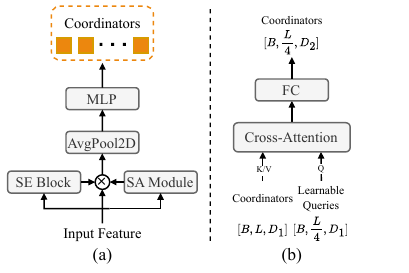} 
\caption{(a) Global Semantic Token Generator. (b) Cross-Attention mechanism in Token Merging block. The notation $[B, L, D_i]$ indicates batch size, sequence length, and feature dimensions.}
\label{fig3}
\end{figure}

\begin{figure*}[t]
\centering
\includegraphics[width=0.95\textwidth]{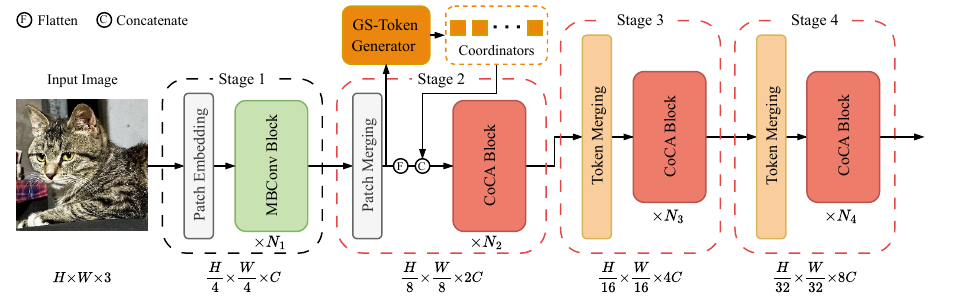}
\caption{Architecture of our proposed CoCAViT. CoCAViT integrates convolutional and transformer-based components in a multi-stage hierarchical structure. The input image is first processed by convolutional patch embedding and MBConv blocks for local feature extraction. Subsequent stages utilize CoCA blocks containing window-based self-attention and global coordinator interaction, enhanced by coordinator tokens produced by a Global Semantic Token Generator.}
\label{fig4}
\end{figure*}

\subsubsection{Domain-Robust Global Semantic Token Generator.}
The foundation of CoCA lies in generating domain-robust global coordinators through our Global Semantic Token Generator (GS-Token Generator), see Figure 3(a). Existing global token initialization in compact Transformers, such as simple pooling or learnable embeddings, often yields domain-sensitive representations—vulnerable to shifts in textures, illumination, or styles—due to inadequate filtering of low-level noise and positional biases. To overcome this, we propose the GS-Token Generator, which extracts global coordinator tokens post-stage 1 downsampling (at 1/8 resolution), providing a stable foundation for subsequent cross-attention in later Transformer stages.

Inspired by domain adaptation insights \cite{ben2010theory}, our generator prioritizes features with enhanced resilience: statistical normalization to decorrelate from domain variations (via implicit LayerNorm in projections), structural and relational cues for semantic stability, and mid-resolution content that balances against extreme frequencies prone to shifts. The architecture operates as follows: starting from input features at 1/8 resolution, we apply Squeeze-and-Excitation (SE) \cite{hu2018squeeze} and Spatial Attention (SA) from CBAM \cite{woo2018cbam} in parallel—the former recalibrates channels to emphasize semantic correlations while downplaying style-sensitive ones, and the latter highlights structural regions over background noise. Outputs are fused via element-wise multiplication, followed by Global Average Pooling (GAP) to aggregate spatial statistics without retaining positional artifacts, and an MLP projection to yield coordinators 
\begin{equation}
\mathbf{G}\in\mathbb{R}^{K\times D}.
\end{equation}
Here, $K=16$ is selected through efficiency-performance trade-offs, while $D$ matches the feature dimension for seamless cross-attention. Executed only once after the convolutional stem, this design incurs minimal overhead yet fosters invariant global patterns, facilitating high mutual information between features and labels ($I(\mathbf{f};\mathbf{y})$ relative to domain indicators $I(\mathbf{f};\mathbf{d})$). By adaptively filtering via SE and SA, the coordinators serve as compressed, generalizable anchors, enabling robust coordination without retraining or added complexity.

\subsubsection{CoCA Mechanism: GGCA + GCWA.}
The CoCA block comprises two sequential components: Gated Global Cross Attention (GGCA) and Global Coordinated Window Attention (GCWA) , as illustrated in Figure 2(a). While window-based self-attention (WSA) in compact Vision Transformers enables efficient local modeling with linear complexity, it inherently creates information silos across windows, limiting global context and rendering models susceptible to domain shifts—such as varying textures or styles that disrupt cross-region coherence. To address this without incurring quadratic overhead, we introduce the Coordinated Cross-Attention (CoCA) mechanism, leveraging a sparse set of coordinators as dynamic global anchors. CoCA establishes a bidirectional information flow: aggregating local patches into enriched coordinators via Gated Global Cross-Attention (GGCA), then broadcasting global insights back to patches through Global-Coordinated Window Attention (GCWA). This design not only bridges window silos but also enhances domain robustness by adaptively filtering and compressing invariant semantic patterns.

Let $\mathbf{P} \in \mathbb{R}^{N \times D}$ denote the flattened patch embeddings (with $N$ patches and dimension $D$), and $\mathbf{G} \in \mathbb{R}^{K \times D}$ the coordinators (with $K \ll N$, typically 16). In GGCA, coordinators both query and contribute to the key-value space alongside patches to aggregate global information, updating themselves while leaving patches unchanged:

\begin{equation}
\mathbf{G}' = \text{Softmax}\left( \frac{\mathbf{Q}_G (\mathbf{K}_G \oplus \mathbf{K}_P)^\top}{\sqrt{D}} \right) (\mathbf{V}_G \oplus \mathbf{V}_P),
\end{equation}
where $\mathbf{Q}_G = \mathbf{G} \mathbf{W}_Q$, $\mathbf{K}_G = \mathbf{G} \mathbf{W}_K$, $\mathbf{V}_G = \mathbf{G} \mathbf{W}_V$, $\mathbf{K}_P = \mathbf{P} \mathbf{W}_K$, and $\mathbf{V}_P = \mathbf{P} \mathbf{W}_V$ are linear projections, with $\oplus$ denoting concatenation. Here, coordinators serve as queries and part of the keys/values, enabling them to retain and enrich their own information while selectively compressing global insights from patches. The softmax induces a gated, sparse attention pattern, allowing each coordinator to focus on semantically relevant patches and self-interactions—fostering freer, more adaptive global modeling than rigid schemes like Swin's shifted windows, which constrain interactions to predefined overlaps.

Subsequently, GCWA integrates this enriched global context into local windows. For each window of patches $\mathbf{P}_w \in \mathbb{R}^{M \times D}$ (with $M$ patches per window), the key-value space is augmented with coordinators:

\begin{equation}
\mathbf{P}_w' = \text{Softmax}\left( \frac{\mathbf{Q}_w (\mathbf{K}_w \oplus \mathbf{K}_G')^\top}{\sqrt{D}} \right) (\mathbf{V}_w \oplus \mathbf{V}_G'),
\end{equation}
where $\mathbf{Q}_w = \mathbf{P}_w \mathbf{W}_Q$, $\mathbf{K}_w = \mathbf{P}_w \mathbf{W}_K$, $\mathbf{V}_w = \mathbf{P}_w \mathbf{W}_V$, $\mathbf{K}_G' = \mathbf{G}' \mathbf{W}_K$, $\mathbf{V}_G' = \mathbf{G}' \mathbf{W}_V$. This unified computation balances local (intra-window) and global (coordinator-derived) signals through learned attention weights, adaptively weighting contributions based on patch needs—e.g., edge patches may rely more on coordinators for cross-window cues.

CoCA forms a complete global-local information processing loop, analyzed from information-theoretic and computational graph perspectives. From an information-theoretic viewpoint, GGCA compresses high-dimensional local patch information ($N \times D$) into a compact global representation ($K \times D$, with $K \ll N$), inherently filtering local noise and domain-specific details to preserve semantically relevant global patterns. The sparse attention mechanism promotes selective information propagation, where different coordinators specialize in distinct semantic regions or modes, reducing redundancy and enhancing representational discriminability. In GCWA, each patch gains contextual enrichment from both local neighbors and global coordinators, enabling adaptive balancing between fine-grained details and overarching semantics for improved comprehension of complex scenes. This setup boosts domain robustness by positioning coordinators as cross-domain semantic anchors, mitigating reliance on domain-specific local features; during domain shifts, these stable global representations provide a corrective foundation for adapting local variations.

From a computational graph lens, CoCA enriches the network structure by disrupting traditional windowed locality with global coordinators, yielding a "locally dense, globally sparse" connectivity pattern that maintains efficiency while facilitating effective global propagation. Unlike methods relying on fixed spatial shifts, CoCA's content-driven attention enables more flexible and adaptive interactions, better suited for handling intricate visual scenarios and domain migration challenges. This principled design promotes hierarchical abstraction—evolving coordinators from mid-level structures to high-level semantics across stages—yielding compact yet resilient representations, as evidenced in later ablations.

\subsubsection{Computational Complexity Analysis.}
CoCA maintains computational efficiency by design. Let $h \times w$ denotes the feature map resolution, $C$ is the channel dimension, $M$ is the window size, $N = hw$ is the number of patches, and $K$ is the number of coordinators (with $K \ll N$). For reference, the complexity of standard multi-head self-attention (MSA) and WSA are:
\begin{equation}
\begin{aligned}
&\Omega(\mathsf{MSA}) = 4hwC^2 + 2(hw)^2C, \\
&\Omega(\mathsf{WSA}) = 4hwC^2 + 2M^2hwC.
\end{aligned}
\end{equation}

For GGCA, coordinators serve as queries and part of the key-value space (concatenated with patches), yielding:

\begin{equation}
\Omega(\mathsf{GGCA}) = 4KC^2 + 2hwC^2 + 2K(K + hw)C.
\end{equation}

This accounts for projections ($4KC^2$ for Q/K/V/O on coordinators, $2hwC^2$ for K/V on patches) and attention computation over the augmented KV space. Likely for GCWA, each window's key-value space is augmented with all K coordinators:

\begin{equation}
\Omega(\mathsf{GCWA}) = 4hwC^2 + 2KC^2 + 2MhwC + 2KhwC.
\end{equation}

A full CoCA (GGCA + GCWA) thus has complexity:

\begin{equation}
\begin{split}
\Omega(\mathsf{CoCA}) &= 6(K+hw)C^2 \\
& \quad + (2K^2+2Mhw+4Khw)C.
\end{split}
\end{equation}

Since $K \ll hw$, terms like $6KC^2$ and $2K^2C$ are negligible, simplifying to $\Omega(\mathsf{CoCA}) \approx 6hwC^2 + 2MhwC + 4KhwC$. This is moderately higher than a single WSA ($\sim 4hwC^2 + 2M^2hwC$) but far more efficient than full MSA, while enabling robust global coordination. The added $4KhwC$ term scales linearly with resolution, preserving compactness and efficiency for larger resolutions.

\begin{table*}[t]
\centering
\small
\label{tab:comparison}
\begin{tabular}{lc|ccc|ccc}
\toprule
Model & Arch. & \makecell{\#Params\\(M)} & \makecell{FLOPs\\(G)} & \makecell{Throughput\\(images/s)} & \makecell{IN-1k\\Top-1(\%)} & \makecell{IN-Real\\Top-1(\%)} & \makecell{IN-V2\\Top-1(\%)} \\
\midrule
LSNet-T & C & 11.4 & 1.5 & 5873 & 74.9 & 81.9 & 62.2 \\
SHViT-S2 & T & 11.5 & 0.4 & 3401 & 75.2 & 82.0 & 62.2 \\
EfficientViT-M5 & T & 12.4 & 0.5 & - & 77.1 & - & - \\
PVTv2-B1 & H & 14.1 & 2.1 & 1242 & 78.7 & 85.2 & 66.9 \\
TinyViT-11M & H & 11.0 & 2.0 & 1325 & 81.5 & 87.1 & 71.1 \\
BiFormer-T & T & 13.1 & 2.2 & 706 & 81.4 & 87.2 & 70.6 \\
TransNeXt-Micro & T & 12.8 & 2.7 & 728 & 82.5 & 87.8 & 72.6 \\
\textbf{CoCAViT-11M(Ours)} & H & 11.4 & 2.2 & 1387 & \textbf{82.7} & \textbf{87.8} & \textbf{72.9} \\
\midrule
DeiT-Small/16 & T & 22.1 & 4.6 & - & 79.9 & - & 68.4 \\
LSNet-B & C & 23.2 & 1.5 & 4587 & 80.3 & 84.1 & 65.2 \\
UniRepLKNet-N & C & 18.3 & 2.8 & - & 81.6 & - & - \\
TinyViT-21M & H & 21.2 & 4.3 & 853 & 83.1 & 88.1 & 73.1 \\
\textbf{CoCAViT-21M(Ours)} & H & 20.6 & 4.1 & 902 & \textbf{83.6} & \textbf{88.3} & \textbf{73.7} \\
\midrule
Swin-T & T & 28.3 & 4.5 & 768 & 81.2 & 86.6 & 69.7 \\
PVTv2-B2 & T & 25.4 & 4.0 & 737 & 82.0 & 87.4 & 71.6 \\
ConvNeXt-T & C & 28.6 & 4.5 & 759 & 82.1 & 87.3 & 71.0 \\
VMamba-T & M & 30.3 & 4.9 & 696 & 82.6 & 87.5 & 72.0 \\
MaxViT-T & T & 30.9 & 5.6 & - & 83.6 & - & - \\
BiFormer-S & T & 25.5 & 4.5 & 414 & 83.8 & 88.3 & 73.6 \\
TransNeXt-Tiny & T & 28.2 & 5.7 & 421 & 84.0 & 88.3 & 73.8 \\
\textbf{CoCAViT-28M(Ours)} & H & 27.8 & 4.9 & 791 & \textbf{84.0} & \textbf{88.4} & \textbf{74.0} \\
\bottomrule
\end{tabular}
\caption{ Comparison of image classification performance on ImageNet-1K and robustness benchmarks. All models are trained and evaluated at 224×224 resolution. Throughput is measured on an RTX 3080 GPU with Intel i7-12700K CPU using batch size 32. Architecture types: C=CNN, T=Transformer, H=Hybrid, M=Mamba. Best accuracy results are highlighted in bold.}
\label{table1}
\end{table*}

\subsection{Architectural Design of CoCAViT}
CoCAViT adopts a hierarchical multi-stage hybrid architecture that progressively builds spatial abstraction while maintaining global coordination, as illustrated in Figure 4. The network begins with convolutional patch embedding followed by MBConv blocks in Stage 1, then transitions to transformer-based stages incorporating CoCA blocks for global-local interaction.

The architecture alternates between WSA blocks and GGCA-GCWA blocks within each stage, as illustrated in Figure 2(b), creating a balanced computation pattern that preserves efficiency while enabling robust global modeling. Each CoCA block incorporates a Convolutional GLU module following TransNeXt~\cite{shi2024transnext}, which enhances feature representation through channel-wise gating and local convolutions. Detailed model configurations are provided in Appendix.

\section{Experiments}
In this section, we conduct a comprehensive evaluation of our CoCAViT models across three different scales to assess their performance and robustness. We perform extensive ablation studies to demonstrate the contribution of each proposed component. For image classification, we evaluate on ImageNet-1K and multiple robustness benchmarks. We also validate the transferability of our approach on downstream tasks including object detection and semantic segmentation.
\subsection{Image Classification}
As our primary evaluation, we train CoCAViT models from scratch on ImageNet-1K and assess their performance on the validation set as well as out-of-distribution benchmarks including ImageNet-Real (IN-Real) \cite{beyer2020we} and MatchedFrequency test set of ImageNet-V2 (IN-V2) \cite{recht2019imagenet}. We adopt the same training recipe as TinyViT \cite{wu2022tinyvit}, with the addition of our anchor loss for coordinator token regularization (detailed hyperparameters provided in the Appendix). All models are trained for 300 epochs on 8 GPUs using automatic mixed precision (AMP) for efficiency.

\subsubsection{Main Results.} Table 1 presents our comparative results against state-of-the-art models across different architectural paradigms, including DeiT\cite{touvron2021training}, Swin Transformer \cite{liu2021swin}, ConvNeXt \cite{liu2022convnet}, PVTv2 \cite{wang2022pvt}, VMamba \cite{liu2024vmamba}, BiFormer \cite{zhu2023biformer}, TinyViT \cite{wu2022tinyvit}, EfficientViT \cite{liu2023efficientvit}, MaxViT \cite{tu2022maxvit}, TransNeXt \cite{shi2024transnext}, SHViT \cite{yun2024shvit}, UniRepLKNet \cite{ding2024unireplknet}, and LSNet \cite{wang2025lsnet}. Our CoCAViT models demonstrate superior performance across all scales while maintaining competitive efficiency. In the compact model range (\~11-14M parameters), CoCAViT-11M achieves 82.7\% top-1 accuracy on ImageNet-1K, substantially outperforming comparable models including TinyViT-11M (81.5\%, +1.2\%), PVTv2-B1 (78.7\%, +4.0\%), and EfficientViT-M5 (77.1\%, +5.6\%). Notably, our model maintains competitive throughput (1387 images/s) while delivering significant accuracy improvements. The performance gains are even more pronounced on robustness benchmarks: 87.8\% on ImageNet-Real (matching TransNeXt-Micro) and 72.9\% on ImageNet-V2, preliminarily demonstrating the effectiveness of our coordinator-based global attention design for robustness retainment. For medium-scale models (\~15-25M parameters), CoCAViT-21M achieves 83.6\% ImageNet-1K accuracy with 20.6 M parameters, outperforming TinyViT-21M (83.1\%, +0.5\%) while using fewer parameters. The model shows consistent improvements on robustness benchmarks with 88.3\% on ImageNet-Real and 73.6\% on ImageNet-V2, indicating strong generalization capabilities. In the larger model category (\~25-30M parameters), CoCAViT-28M reaches 84.0\% ImageNet-1K accuracy, matching TransNeXt-Tiny while using fewer parameters (27.8M vs. 28.2M). More importantly, our model achieves superior throughput (791 vs. 421 images/s) and maintains the highest robustness performance with 88.4\% on ImageNet-Real and 74.0\% on ImageNet-V2.

\subsubsection{Efficiency Analysis.} Despite incorporating coordinator-based cross-attention for enhanced global modeling, CoCAViT achieves favorable speed-accuracy trade-offs across all model scales: CoCAViT-11M delivers 1387 images/s while significantly outperforming faster models like LSNet-T (5873 images/s, 74.9\% accuracy) and achieving comparable speed to TinyViT-11M (1325 images/s) with superior accuracy. This efficiency stems from our careful design choices, including the use of coordinator tokens that scale with window numbers rather than total patch count, and the progressive MLP ratio strategy that reduces computational overhead in deeper layers. Taking advantages from the hybrid architecture and window attention mechanism, our models also demonstrate clear computational advantages over high-performing TransNeXt models. For example, CoCAViT-28M achieves ImageNet-1K accuracy (84. 0\%) comparable to TransNeXt-Tiny while delivering nearly 2× higher throughput (791 vs 421 images/s) and using fewer FLOPs (4.9 G vs 5.7 G). This efficiency stems from our principled design choices: coordinator tokens scale with window numbers rather than total patch count, the progressive MLP ratio reduces computational overhead in deeper layers, and the hybrid CNN-Transformer architecture leverages hardware-optimized convolution operations in early stages.

\begin{table}[h]
\centering
\begin{tabular}{l|cccc}
\toprule
Model & \makecell{A} & \makecell{R} & \makecell{Sketch} \\
\midrule
Swin-T & 21.1 & 41.8 &  29.3 \\
ConvNeXt-T & 24.2 & 47.6 &  33.8 \\
VMamba-T & 27.0 & 45.9 &  32.9 \\
TinyViT-11M & 26.0 & 44.8 &  - \\
TinyViT-21M & 34.9 & 49.2 &  - \\
TransNeXt-Micro & 29.9 & 46.2 & 33.0 \\
TransNeXt-Tiny & 39.3 & 50.2 & - \\
\midrule
CoCAViT-11M(Ours) & 34.1 & 47.8 & 35.5 \\
CoCAViT-21M(Ours) & 38.9 & 50.1 & 39.6 \\
\textbf{CoCAViT-28M(Ours)} & \textbf{39.8} & \textbf{51.1} &  \textbf{40.2} \\
\bottomrule
\end{tabular}
\caption{Robustness comparisons of different models. All models are trained and evaluated at 224×224 resolution. Throughput is measured on an RTX 3080 GPU with Intel i7-12700K CPU using batch size 32. Architecture types: C=CNN, T=Transformer, H=Hybrid, M=Mamba. Best accuracy results are highlighted in bold.}
\label{table4}
\end{table}

\subsubsection{Domain Generalization Evaluation.} To further assess the generalization capabilities of our models, we conduct extensive evaluation on multiple domain shift and robustness benchmarks, including ImageNet-R \cite{hendrycks2021many}, ImageNet-A \cite{hendrycks2021natural}, and ImageNet-Sketch \cite{wang2019learning}. As shown in Table 4, our CoCAViT models demonstrate remarkable robustness across all scales, with particularly impressive performance in the compact model regime where existing methods suffer disproportionate degradation.

The results validate a key architectural insight: our coordinator-based cross-attention mechanism effectively bridges the robustness gap that typically plagues efficient models. CoCAViT-11M achieves 34.1\% on ImageNet-A and 47.8\% on ImageNet-R, substantially outperforming comparable compact models like Swin-T (21.1\%/41.8\%) and ConvNeXt-T (24.2\%/47.6\%). More remarkably, our smallest model maintains generalization performance competitive with much larger architectures—CoCAViT-11M's ImageNet-Sketch accuracy (35.5\%) even surpasses VMamba-T (32.9\%) despite using significantly fewer parameters.

This robustness stems from our principled structural innovations: the hybrid CNN-Transformer design provides stable low-level features through convolutional stages, while our CoCA mechanism restores global coordination without the brittleness of purely local attention. The coordinator tokens act as adaptive semantic anchors, enabling the model to maintain coherent global understanding even under severe distribution shifts. Notably, CoCAViT-28M achieves 39.8\% on ImageNet-A and 51.1\% on ImageNet-R, approaching the performance of specialized large-scale robust models while maintaining the efficiency of compact architectures. These results demonstrate that thoughtful architectural design can fundamentally enhance robustness in efficient models, rather than merely trading off accuracy for speed.

\section{Conclusion}
In this work, we address a critical yet underexplored challenge in efficient vision transformers: the robustness-efficiency gap that plagues current lightweight architectures when facing distribution shifts. We propose CoCAViT framework to confront these challenges through principled architectural innovations. The implications extend beyond immediate performance gains; they point to a more principled approach to efficient model design where robustness is engineered from the ground up, not bolted on as an afterthought.

\appendix
\begin{table*}[h]
\centering
\small
\begin{minipage}[h]{0.65\textwidth}
\centering
\begin{tabular}{l|cccccc}
\toprule
Backbone & \makecell{$\mathrm{AP}^{box}$} & \makecell{$\mathrm{AP}^{box}_{50}$} & \makecell{$\mathrm{AP}^{box}_{75}$} & \makecell{$\mathrm{AP}^{mask}$} & \makecell{$\mathrm{AP}^{mask}_{50}$} & \makecell{$\mathrm{AP}^{mask}_{75}$} \\
\midrule
ConvNeXt-T & 50.4 & 69.1 & 54.8 & 43.7 & 66.5 & 47.3 \\
Swin-T & 50.4 & 69.2 & 54.7 & 43.7 & 66.6 & 47.3 \\
PVTv2-B2 & 51.1 & 69.8 & 55.3 & - & - & - \\
MaxViT-T & 52.1 & \textbf{71.9} & 56.8 & 44.6 & \textbf{69.1} & 48.4 \\
\midrule
CoCAViT-21M(Ours) & 51.8 & 70.5 & 56.1 & 44.9 & 67.8 & 48.2 \\
\textbf{CoCAViT-28M(Ours)} & \textbf{52.2} & 71.0 & \textbf{56.8} & \textbf{45.2} & 68.9 & \textbf{48.8} \\
\bottomrule
\end{tabular}
\caption{Two-stage object detection \& instance segmentation results on COCO 2017 using Cascade Mask-RCNN framework. Best accuracy results are highlighted in bold.}
\label{table2}
\end{minipage}
\begin{minipage}[h]{0.32\textwidth}
\centering
\begin{tabular}{l|c}
\toprule
Backbone & \makecell{mIoU} \\
\midrule
Swin-T & 45.8 \\
ConvNeXt-T & 46.7 \\
VMamba-T & 48.8 \\
BiFormer-S & 50.8 \\
\midrule
CoCAViT-21M(Ours) & 50.8 \\
\textbf{CoCAViT-28M(Ours)} & \textbf{51.3} \\
\bottomrule
\end{tabular}
\captionsetup{width=0.85\textwidth}
\caption{Semantic segmentation results on ADE20K using UperNet framework. Results are reported with multi-scale testing.}
\label{table3}
\end{minipage}

\end{table*}
\section{A. Configurations of CoCAViT Variants}
\begin{table}[h]
\centering
\small
\begin{tabular}{l|cccc}
\toprule
Size & 11M & 21M & 28M \\
\midrule
conv dims & 72 & 96 & 96 \\
head dims & 24 & 24 & 24 \\
heads num. & [-, 4, 8, 14] & [-, 6, 12, 18] & [-, 6, 12, 18] \\
depths & [2, 2, 12, 2] & [2, 2, 12, 2] & [2, 2, 15, 2] \\
MLP ratio & [-, 5, 4, 3] & [-, 5, 4, 3] & [-, 5, 4, 3] \\
window size & [-, 7, 7, 7] & [-, 7, 7, 7] & [-, 7, 7, 7] \\
interaction & [2, 3, -1] & [2, 3, -1] & [2, 3, -1] \\ 
\bottomrule
\end{tabular}
\label{tab:cocavit-config}
\caption{The configurations of CoCAViT variants for 224 × 224 resolution. Interaction means the frequency of using GGCA + GCWA, otherwise WSA.}
\end{table}

\section{B. Ablation Studies}
To understand the contribution of each proposed component, we conduct brief ablation studies examining the impact of different backbone reconfiguration strategies, as well as the effectiveness of the CoCA mechanism, demonstrating that each component contributes meaningfully to the final performance.
\begin{table}[h]
\centering
\small
\begin{tabular}{l|c|ccc}
\toprule
Step & Method & \#Params.(M) & IN-1K Top1(\%) \\
\midrule
0 & Swin-Tiny & 28.3 & 81.2 \\
1 & pure WSA & 28.3 & 80.1(-1.1) \\
2 & MBConv Stage & 18.9 &  80.6(+0.5) \\
3 & CoCA Block & 21.3 & 82.1(+1.5) \\
4 & \makecell{wider conv\\deeper transformer} & 20.1 & 83.1(+1.0) \\
5 & Convolutional GLU & 20.6 & 83.6(+0.5) \\
\bottomrule
\end{tabular}
\label{tab:cocavit-ablation}
\caption{The ablation experiments demonstrating the roadmap from Swin Transformer-Tiny to CoCAViT-21M.}
\end{table}

\section{C. Downstream Tasks}
\subsubsection{Object Detection and Instance Segmentation.} We employ Cascade Mask-RCNN \cite{cai2019cascade} as the detection head and evaluate on COCO 2017 \cite{lin2014microsoft} following the experimental setup of Swin Transformer \cite{liu2021swin}. The backbone networks are fine-tuned for 36 epochs using the 3× training schedule. As presented in Table 2, our models demonstrate consistent improvements across all detection and segmentation metrics. CoCAViT-21M achieves 51.8 $AP^{box}$ and 44.9 $AP^{mask}$ , outperforming comparable baselines including PVTv2-B2 (51.1/43.7) and Swin-T (50.4/43.7). CoCAViT-28M further improves to 52.2 $AP^{box}$ and 45.2 $AP^{mask}$ , approaching the performance of the larger MaxViT-T model (52.1/44.6) while using fewer parameters.

\subsubsection{Semantic Segmentation.} We adopt the UperNet \cite{xiao2018unified} framework and evaluate on ADE20K \cite{zhou2017scene} following Swin Transformer's setup. Models are trained for 160k iterations. As shown in Table 3, our models demonstrate strong performance improvements over baseline architectures. CoCAViT-21M achieves 50.8 mIoU, matching the performance of BiFormer-S while using fewer parameters (20.6M vs. 25.5M). CoCAViT-28M further improves to 51.2 mIoU, outperforming all compared baselines including the strong BiFormer-S by +0.4 points.

The performance gains in object detection is particularly notable in high-precision regimes (AP75), where CoCAViT-28M achieves 56.8 $AP_{75}^{box}$ compared to Swin-T's 54.7, indicating improved localization accuracy. We believe these improvements stem from the enhanced global context modeling provided by our coordinator mechanism, which enables better understanding of object relationships, scene-level context, and long-range spatial dependencies—crucial for accurate detection in cluttered scenes and dense prediction tasks like semantic segmentation. The global coordination helps distinguish similar objects, provides robust feature representations that generalize well from ImageNet pretraining to complex multi-object scenarios, and allows better integration of multiscale features for accurate boundary delineation. The consistent gains across model scales and tasks validate that our coordination mechanism effectively transfers pretraining benefits to downstream applications.

\section{D. Anchor Loss for Robust Coordination}
To ensure coordinators learn diverse, domain-invariant representations, we introduce an anchor loss that promotes orthogonality and stability:
\begin{equation}
\mathcal{L}_{\mathrm{anchor}}=0.5\cdot\mathcal{L}_{\mathrm{diversity}}+0.1\cdot\mathcal{L}_{\mathrm{stability}},
\end{equation}
where the diversity term encourages coordinator orthogonality:
\begin{equation}
\mathcal{L}_{\mathrm{diversity}}=\|\mathbf{G}^T\mathbf{G}-\mathbf{I}\|_F^2,
\end{equation}
and the stability term prevents coordinator collapse:
\begin{equation}
\mathcal{L}_{\mathrm{stability}}=\|\mathbf{G}-\bar{\mathbf{G}}\|_{F}^{2},
\end{equation}
with $\bar{\mathbf{G}}$ being the batch-wise coordinator mean. This regularization ensures that coordinators maintain distinct, stable global patterns that generalize across domains.

\bibliography{aaai2026}

\end{document}